\documentclass{ARKkapproc}
\usepackage{epsfig}
\usepackage{subeqn}
 \kluwerbib
\newcommand{\gbf}[1] {\mbox{\boldmath${#1}$\unboldmath}}
\newcommand{\be}{\begin{equation}}
\newcommand{\ee}{\end{equation}}
\newcommand{\beq}{\begin{equation}}
\newcommand{\eeq}{\end{equation}}
\newcommand{\bed}{\begin{displaymath}}
\newcommand{\eed}{\end{displaymath}}
\newcommand{\beqa}{\begin{eqnarray}}
\newcommand{\eeqa}{\end{eqnarray}}
\newcommand{\beqann}{\begin{eqnarray*}}
\newcommand{\eeqann}{\end{eqnarray*}}
\newcommand{\bseq}{\begin{subequations}}
\newcommand{\eseq}{\end{subequations}}

\newcommand{\ba}{\begin{array}}
\newcommand{\ea}{\end{array}}

\newcommand{\negr}[1]{{\bf {#1}}}

\begin{document}
\articletitle{Workspace Analysis \goodbreak of the  Orthoglide
\goodbreak using Interval Analysis}
\author{D. Chablat, Ph. Wenger}
 \affil{Institut de Recherche en Communications et Cybern\'etique
de Nantes,
 1, rue de la No\"e, 44321 Nantes, France \\
\email{Damien.Chablat\symbol{64}irccyn.ec-nantes.fr}}
\author{J. Merlet}
 \affil{INRIA Sophia-Antipolis,
  2004 Route des Lucioles, 06902 Sophia Antipolis, France \\
  \email{Merlet\symbol{64}sophia.inria.fr}}
\begin{abstract}
This paper addresses the workspace analysis of the orthoglide, a
3-DOF parallel mechanism designed for machining applications. This
machine features three fixed parallel linear joints which are
mounted orthogonally and a mobile platform which moves in the
Cartesian $x$-$y$-$z$ space with fixed orientation. The workspace
analysis is conducted on the bases of prescribed kinetostatic
performances. The interesting features of the orthoglide are a
regular Cartesian workspace shape, uniform performances in all
directions and good compactness. Interval analysis based methods
for computing the dextrous workspace and the largest cube enclosed
in this workspace are presented.
\end{abstract}
\section{Introduction}
Parallel kinematic machines (PKM) are commonly claimed to offer
several advantages over their serial counterparts, like high
structural rigidity, high dynamic capacities and high accuracy
(\cite{Treib,Wenger_a}). Thus, PKM are interesting alternative
designs for high-speed machining applications.

This is why parallel kinematic machine-tools attract the interest
of more and more researchers and companies. Since the first
prototype presented in 1994 during the IMTS in Chicago by
Gidding\&Lewis (the Variax), many other prototypes have appeared.

However, the existing PKM suffer from two major drawbacks, namely,
a complex Cartesian workspace and highly non linear input/output
relations. For most PKM, the Jacobian matrix which relates the
joint rates to the output velocities is not constant and not
isotropic. Consequently, the performances (e.g. maximum speeds,
forces accuracy and rigidity) vary considerably for different
points in the Cartesian workspace and for different directions at
one given point. This is a serious drawback for machining
applications (\cite{Kim,Treib,Wenger_b}). To be of interest for
machining applications, a PKM should preserve good workspace
properties, that is, regular shape and acceptable kinetostatic
performances throughout. In milling applications, the machining
conditions must remain constant along the whole tool path
(\cite{Rehsteiner_a,Rehsteiner_b}). In many research papers, this
criterion is not taking into account in the algorithmic methods
used for the optimization of the workspace volume
(\cite{Luh,Merlet, Ottaviano}).

The orthoglide is a $3$-axis PKM with the advantages a classical
serial PPP machine tool but not its drawbacks. It is an optimized
version of the Delta mechanism defined by \cite{Clavel:1990}. Most
industrial 3-axis machine-tool have a serial PPP kinematic
architecture with orthogonal linear joint axes along the x, y and
z directions. Thus, the motion of the tool in any of these
directions is linearly related to the motion of one of the three
actuated axes. Also, the performances are constant in the most
part of the Cartesian workspace, which is a parallelepiped. The
main drawback is inherent to the serial arrangement of the links,
namely, poor dynamic performances.

The orthoglide is a PKM with three fixed linear joints mounted
orthogonally. The mobile platform is connected to the linear
joints by three articulated parallelograms and moves in the
Cartesian x-y-z space with fixed orientation. Its workspace shape
is close to a cube whose sides are parallel to the planes $xy$,
$yz$ and $xz$ respectively. The interval analysis is conducted on
the basis of prescribed bounded velocity and force transmission
factors. Interval analysis based method is used to compute
dextrous workspace as well as the largest cube enclosed in this
workspace (\cite{Merlett}).

\section{Description of the Orthoglide}
Most existing PKM can be classified into two main families. The
PKM of the first family have fixed foot points and variable length
struts and are generally called ``hexapods''. The second family of
PKM has been more recently investigated and have variable foot
points and fixed length struts. PKMs of the second family are more
interesting because the actuators are fixed and thus the moving
masses are lower than in the hexapods and tripods.

The  orthoglide studied in this article is a $3$-axis
translational parallel kinematic machine and is belongs to the
second family. Figure~\ref{figure:Orthoglide} shows the general
kinematic architecture of the orthoglide. The orthoglide has three
parallel $PRPaR$ identical chains (where $P$, $R$ and $Pa$ stands
for Prismatic, Revolute and Parallelogram joint, respectively).
The actuated joints are the three orthogonal linear joints. These
joints can be actuated by means of linear motors or by
conventional rotary motors with ball screws. The output body is
connected to the linear joints through a set of three
parallelograms of equal lengths $L~=~B_iC_i$, so that it can move
only in translation. The first linear joint axis is parallel to
the $x$-axis, the second one is parallel to the $y$-axis and the
third one is parallel to the $z$-axis. In
figure~\ref{figure:Orthoglide}, the base points $A_1$, $A_2$ and
$A_3$ are fixed on the $i^{th}$ linear axis such that
$A_1A_2=~A_1A_3=~A_2A_3$, $B_i$ is at the intersection of the
first revolute axis $\negr i_i$ and the second revolute axis
$\negr j_i$ of the $i^{th}$ parallelogram, and $C_i$ is at the
intersection of the last two revolute joints of the $i^{th}$
parallelogram. When each $B_iC_i$ is aligned with the linear joint
axis $A_iB_i$ , the orthoglide is in an isotropic configuration
and the tool center point $P$ is located at the intersection of
the three linear joint axes. In this configuration, the base
points $A_1$, $A_2$ and $A_3$ are equally distant from $P$. The
symmetric design and the simplicity of the kinematic chains (all
joints have only one degree of freedom,
fig.~\ref{figure:Leg_Kinematic}) should contribute to lower the
manufacturing cost of the orthoglide. The orthoglide is free of
singularities and self-collisions. The workspace has a regular,
quasi-cubic shape. The input/output equations are simple and the
velocity transmission factors are equal to one along the $x$, $y$
and $z$ direction at the isotropic configuration, like in a serial
$PPP$ machine (\cite{Wenger}).

\section{Kinematic Equations and Singularity Analysis}
We recall briefly here the kinematics of the Orthoglide (See \cite{Wenger} for more details).
\subsection{Kinematic Equations}
\begin{figure}[!hb]
    \begin{center}
    \begin{tabular}{cc}
       \begin{minipage}[t]{65 mm}
           \centerline{\hbox{\includegraphics[width=54mm,height=38mm]{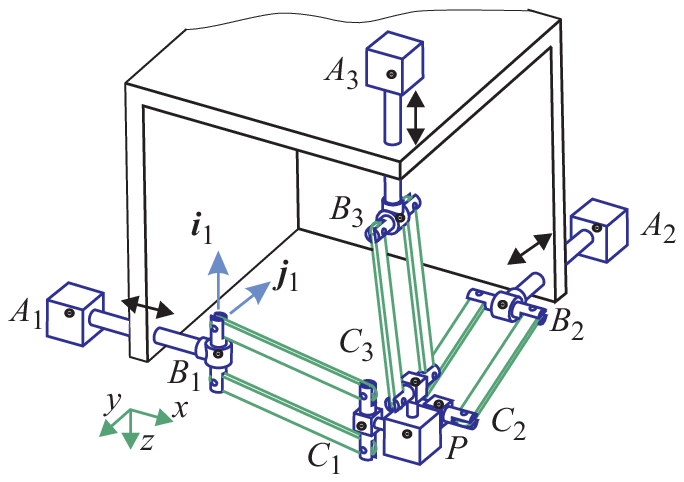}}}
           \caption{Orthoglide kinematic architecture}
           \protect\label{figure:Orthoglide}
       \end{minipage} &
       \begin{minipage}[t]{55 mm}
           \centerline{\hbox{\includegraphics
           [width=48mm,height=38mm]{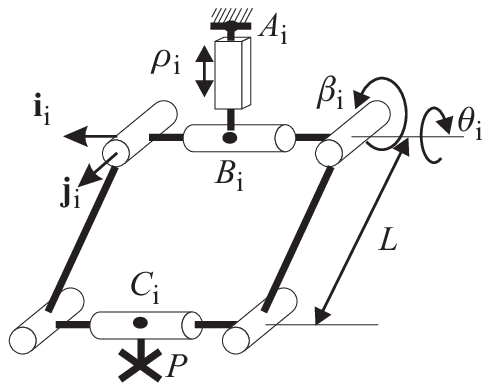}}}
           \caption{Leg kinematics}
           \protect\label{figure:Leg_Kinematic}
       \end{minipage}
    \end{tabular}
    \end{center}
\end{figure}

Let $\theta_i$ and $\beta_i$ denote the joint angles of the parallelogram about the axes $\negr i_i$ and $\negr j_i$, respectively (fig.~\ref{figure:Leg_Kinematic}). Let $\negr \rho_1$, $\negr \rho_2$, $\negr \rho_3$ denote the linear joint variables, $\negr \rho_i=A_iB_i$ and $L$ denote the length of the tree legs, $B_iC_i$. The position vector \negr p of the tool center point $P$ is defined in a reference frame (O, $x$, $y$, $z$) centered at the intersection of the three linear joint axes (note that the reference frame has been translated in Fig.~\ref{figure:Orthoglide} for more legibility).

Let $\dot{\gbf {\rho}}$ be referred to as the vector of actuated joint rates and $\dot{\negr p}$ as the velocity vector of point $P$:
 \bed
    \dot{\gbf{\rho}}=
    [\dot{\rho}_1~\dot{\rho}_2~\dot{\rho}_3]^T
   ,\quad
    \dot{\negr p}=
    [\dot{x}~\dot{y}~\dot{z}]^T
 \eed
$\dot{\negr p}$ can be written in three different ways by
traversing the three chains $A_iB_iC_iP$:
 \bseq
 \beqa
    \dot{\negr p} \!\!\!&=&\!\!\!\!
    \negr n_1 \dot{\rho}_1 +
    (\dot{\theta}_1 \negr i_1 + \dot{\beta}_1 \negr j_1)
    \times
    (\negr c_1 - \negr b_1) \\
    \dot{\negr p} \!\!\!&=&\!\!\!\!
    \negr n_2 \dot{\rho}_1 +
    (\dot{\theta}_2 \negr i_2 + \dot{\beta}_2 \negr j_2)
    \times
    (\negr c_2 - \negr b_2) \\
    \dot{\negr p} \!\!\!&=&\!\!\!\!
    \negr n_3 \dot{\rho}_3 +
    (\dot{\theta}_3 \negr i_3 + \dot{\beta}_3 \negr j_3)
    \times
    (\negr c_3 - \negr b_3)
 \eeqa
 \label{equation:cinematique}
 \eseq
where $\negr b_i$ and $\negr c_i$ are the position vectors of the
points $B_i$ and $C_i$, respectively, and $\negr n_i$ is the
direction vector of the linear joints, for $i=1, 2, $3.
\subsection{Singular configurations}
We want to eliminate the two idle joint rates $\dot{\theta}_i$ and
$\dot{\beta}_i$ from Eqs.~(\ref{equation:cinematique}a--c), which
we do upon dot-multiplying Eqs.~(\ref{equation:cinematique}a--c)
by $\negr c_i - \negr b_i$:
 \bseq
 \beqa
   (\negr c_1 - \negr b_1)^T \dot{\negr p} &=&
   (\negr c_1 - \negr b_1)^T
   \negr n_1 \dot{\rho}_1 \\
   (\negr c_2 - \negr b_2)^T \dot{\negr p} &=&
   (\negr c_2 - \negr b_2)^T
   \negr n_2 \dot{\rho}_2 \\
   (\negr c_3 - \negr b_3)^T \dot{\negr p} &=&
   (\negr c_3 - \negr b_3)^T
   \negr n_3\dot{\rho}_3
 \eeqa
 \label{equation:cinematique-2}
 \eseq
Equations (\ref{equation:cinematique-2}a--c) can now be cast in
vector form, namely
 \bed
   \negr A \dot{\bf p} = \negr B \dot{\gbf \rho}
 \eed
where \negr A and \negr B are the parallel and serial Jacobian
matrices, respectively:
 \beqa
   \negr A =
   \left[\begin{array}{c}
           (\negr c_1 - \negr b_1)^T \\
           (\negr c_2 - \negr b_2)^T \\
           (\negr c_3 - \negr b_3)^T
         \end{array}
   \right]
   {\rm ~~and~~}
   \negr B =
   \left[\begin{array}{ccc}
            \eta_1&
            0 &
            0 \\
            0 &
            \eta_2&
            0 \\
            0 &
            0 &
            \eta_3
         \end{array}
   \right]
 \label{equation:A_et_B}
 \eeqa
with $\eta_i= (\negr c_i - \negr b_i)^T \negr n_i $ for $i=1,2,3$.
\par
Parallel singularities (\cite{Chablat}) occur when the determinant
of the matrix \negr A vanishes, {\it i.e.} when $det(\negr A)=0$.
In such configurations, it is possible to move locally the mobile
platform whereas the actuated joints are locked. These
singularities are particularly undesirable because the structure
cannot resist any force. Eq.~(\ref{equation:A_et_B}a) shows that
the parallel singularities occur when:
 \bed
    (\negr c_1 - \negr b_1) =
    \alpha  (\negr c_2 - \negr b_2) +
    \lambda (\negr c_3 - \negr b_3)
 \eed
that is when the points $B_1$, $C_1$, $B_2$, $C_2$, $B_3$ and
$C_3$ are coplanar. A particular case occurs when the links
$B_iC_i$ are parallel:
 \bed
    (\negr c_1 - \negr b_1) //
    (\negr c_2 - \negr b_2)
    \quad {\rm and} \quad 
    (\negr c_2 - \negr b_2) //
    (\negr c_3 - \negr b_3)
    \quad {\rm and} \quad 
    (\negr c_3 - \negr b_3) //
    (\negr c_1 - \negr b_1)
 \eed
Serial singularities arise when the serial Jacobian matrix \negr B
is no longer invertible {\it i.e.} when $det(\negr B)=0$. At a
serial singularity a direction exists along which any cartesian
velocity cannot be produced. Eq.~(\ref{equation:A_et_B}b) shows
that $\det(\negr B)=0$ when for one leg $i$, $(\negr b_i - \negr
a_i) \perp (\negr c_i - \negr b_i)$.

When \negr A and \negr B are not singular, we obtain the
relations,
  \be
   \dot{\negr p} = \negr J  \dot{\gbf \rho} {\rm ~with~}
   \negr J = \negr A^{-1} \negr B
  \ee
\subsection{Velocity transmission factors}
For joint rates belonging to a unit ball, namely, $||\dot{\gbf
\rho}|| \leq 1$, the Cartesian velocities belong to an ellipsoid
such that:
 \bed
   \dot{\gbf p}^T (\negr J \negr J^T) \dot{\gbf p} \leq 1
 \eed
The eigenvectors of matrix $\negr J \negr J^T$ define the
direction of its principal axes of this ellipsoid and the square
roots $\psi_1$, $\psi_2$ and $\psi_3$ of the eigenvalues
$\sigma_1$, $\sigma_2$ and $\sigma_3$ of $\negr J \negr J^T$, {\it
i.e.} the lengths of the aforementioned principal axes are the
velocity transmission factors in the directions. To limit the
variations of this factor in the Cartesian workspace, we set
 \beqa
   \psi_{min} \leq \psi_i \leq \psi_{max}
   \label{e:velocity_limits}
 \eeqa
throughout the workspace.
To simplify the problem, we set $\psi_{min}=1/ \psi_{max}$ where
the value of $\psi_{max}$ depends on the performance requirements.
\section{Determination of the dextrous workspace}
The dextrous workspace ${\cal W}$ is here defined as the loci of
the points for which all the eigenvalues  of the matrix $\negr J
\negr J^T$, {\it i.e.} the velocity transmission factors, lie
within a predefined range $[\sigma_{min},\sigma_{max}]$. These
eigenvalues are determined by solving the third degree
characteristic polynomial of the matrix which is defined only for
the points within the intersection ${\cal I}$ of the three
cylinders defined by
 \be
 x^2+y^2 \le L \quad x^2+z^2 \le L \quad y^2+z^2 \le L
 \ee
To solve numerically the above equations, the length of the legs
is normalized, {\it i.e.} we set $L=1$.

Our purpose in this section is to determine an approximation of
${\cal W}$ as a set of 3D Cartesian boxes for any point of which
we are sure that the constraints on the eigenvalues are satisfied.
The width of all the boxes in the list will be greater than a
given threshold and the value of this threshold will define the
quality of the approximation.
\subsection{Box verification}
A basic tool of the algorithm is a module ${\cal M}(B)$ that takes
as input a box $B$ belonging to ${\cal I}$ and whose output is:
\begin{itemize}
\item either that for any point in the box the eigenvalues lie in the
range $[\sigma_{min},\sigma_{max}]$
\item or that for any point in the box one of the eigenvalues is either
lower than $\sigma_{min}$ or larger than $\sigma_{max}$
\item or that the two previous conditions does not hold for all the
points of the box {\it i.e.} that for some points the eigenvalues
lie in the range $[\sigma_{min},\sigma_{max}]$ while this is not
true for some other points
\end{itemize}
The first step of this module consists in considering an arbitrary
point of the box (e.g. its center) and to compute the eigenvalues
at this point: either all of them lie in the range
$[\sigma_{min},\sigma_{max}]$ or at least one of them lie outside
this range.

In the first case if we are able to check that there is no point
in $B$ such that the eigenvalue at this point may be equal to
$\sigma_{min}$ or $\sigma_{max}$, then we may guarantee that for
any point of $B$ the eigenvalues will be in the range
$[\sigma_{min},\sigma_{max}]$. Indeed assume that at a given point
of $B$ the lowest eigenvalue is lower than $\sigma_{min}$: this
implies that somewhere along the line joining this point to the
center of the box the lowest eigenvalue will be exactly
$\sigma_{min}$. To perform this check we substitute in the
polynomial $\lambda$ successively by $\sigma_{min}$ and
$\sigma_{max}$ to get a polynomial in $x, y, z$ only. We have now
to verify if there is at least one value for these three variables
that cancel the polynomial, being understood that these values
have to define a point belonging to $B$: this is done by using an
interval analysis algorithm from the {\tt ALIAS}
library~(\cite{Merlett}).

Assume now that at the center of the box the largest eigenvalue is
greater than $\sigma_{max}$. If we are to determine that there is
no point of $B$ such that one of the eigenvalue is equal to
$\sigma_{max}$, then we may guarantee that for any point of $B$
the largest eigenvalue will always be greater than $\sigma_{max}$.
This check is performed by using the same method than in the
previous case. Hence the ${\cal M}$ module will return:
\begin{itemize}
\item 1: if for all points of $B$ the eigenvalues lie in
$[\sigma_{min},\sigma_{max}]$ (hence $B$ is in the dextrous
workspace)
\item -1: if for all points of $B$ either the largest eigenvalue is
always greater than $\sigma_{max}$ or the lowest eigenvalue is
lower than $\sigma_{min}$ (hence $B$ is outside the dextrous
workspace).
\item 0: in the other cases {\it i.e.} parts of $B$ may be either outside or
inside the dextrous workspace
\end{itemize}
\subsection{Algorithm for the determination of the dextrous workspace}
The principle of the algorithm is pretty simple: we will maintain
a list ${\cal L}$ of boxes, indexed by $i$, which is is
initialized with the box [-1,1], [-1,1], [-1,1]. A minimal width
$\epsilon$ for the ranges in a box is defined and the operator
${\cal W}(B_i)$  will return the largest width of the ranges in
$B_i$. An error index ${\cal E}$ will be computed as the total
volume of the boxes that are not in the approximation but may
contain points that are inside the dextrous workspace. We then
apply ${\cal M}(B_i)$:
\begin{itemize}
\item if ${\cal M}(B_i)$=1: we store $B_i$ as part of the dextrous
workspace and consider the next box in ${\cal L}$
\item if ${\cal M}(B_i)$=-1: we consider the next box in ${\cal L}$
\item if ${\cal M}(B_i)$=0:
        \begin{itemize}
        \item if ${\cal W}(B_i) \ge \epsilon$: we create 2 new boxes
from $B_i$ by bisecting the range of $B_i$ with the largest width.
The two new boxes are put at the end of ${\cal L}$
        \item otherwise we add the volume of $B_i$ to ${\cal E}$
        \end{itemize}
\end{itemize}
The algorithm stops when all the boxes in ${\cal L}$ have been
processed. Note that this basic algorithm has to be modified in
order to consider only boxes that belongs to ${\cal I}$ but this
can be done using the same principle. The algorithm returns a
description of the dextrous workspace as a list of boxes and the
comparison between the volume of the approximation and ${\cal E}$
allows to determine the quality of the approximation.
 Note that ${\cal E}$ is
very conservative as part of this volume consists in points that
do not belong to ${\cal I}$ or to the dextrous workspace.
\subsection{Implementation and results}
The previous algorithms has been implemented in Maple with a
system call to a C++ program that implements the ${\cal M}$
module. For an $\epsilon$ of 0.05 we found in about 5 hours that
the volume of the dextrous workspace is 1.468 with an error bound
of [0,0.48] with $\sigma_{min}=0.25$ and $\sigma_{max}=4$, {\it
i.e.} $\psi_{min}=1/2$ and $\psi{max}=2$
(Fig.~\ref{figure:Dextrous_Configuration}).
\section{Determination of the largest cube enclosed in the dextrous
workspace}
\begin{figure}[hbt]
    \begin{center}
    \begin{tabular}{cc}
       \begin{minipage}[t]{45 mm}
           \centerline{\hbox{\includegraphics[width=45mm,height=45mm]{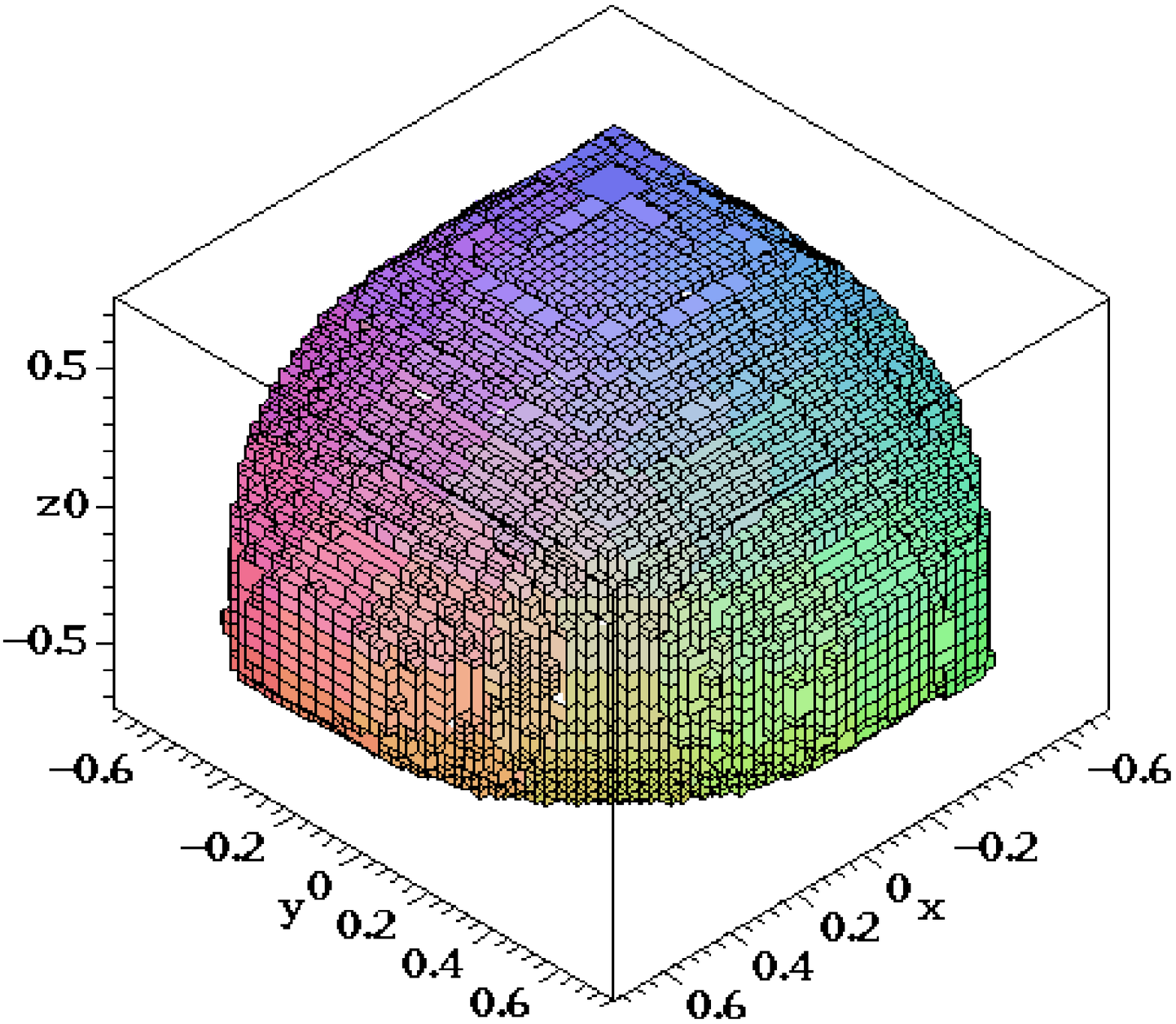}}}
           \caption{Dextrous workspace of the Orthoglide mechanism}
           \protect\label{figure:Dextrous_Configuration}
       \end{minipage} &
       \begin{minipage}[t]{65 mm}
           \centerline{\hbox{\includegraphics[width=55mm,height=55mm]{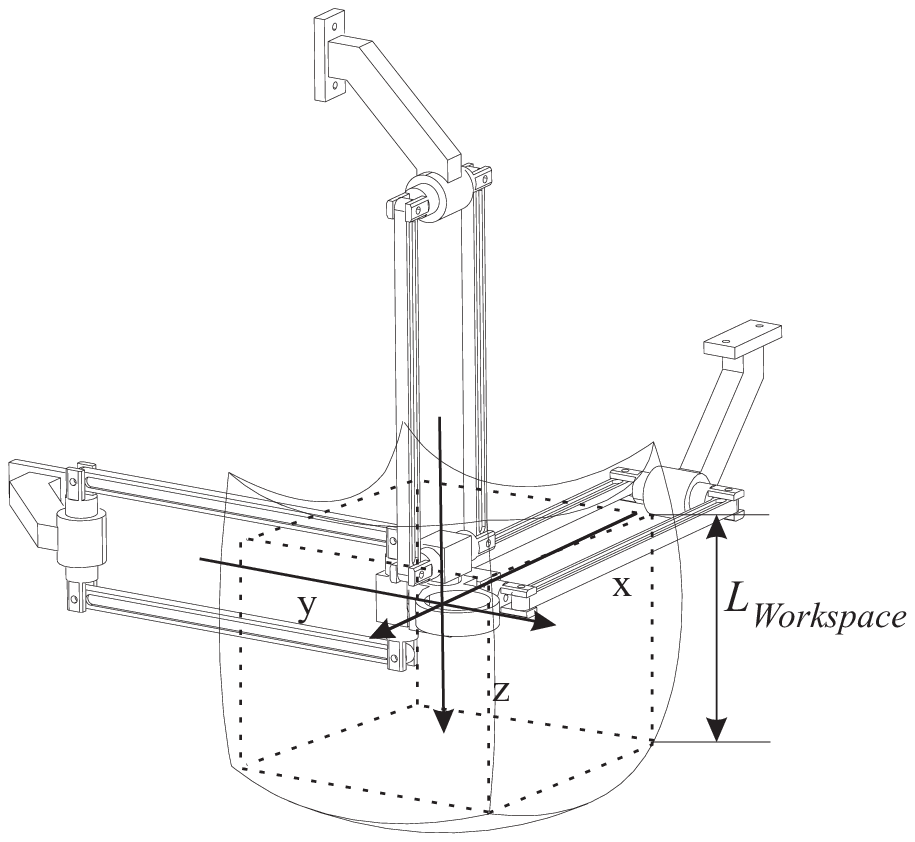}}}
           \caption{Cartesian workspace and isotropic configuration of the Orthoglide mechanism}
           \protect\label{figure:Orthoglide_Workspace_Isotropic_Configuration}
       \end{minipage}
    \end{tabular}
    \end{center}
\end{figure}

For usual machine tools, the Cartesian workspace is generally
given as a function of the size of a right-angled parallelepiped.
Due to the symmetrical architecture of the orthoglide, the
Cartesian workspace has a fairly regular shape in which it is
possible to include a cube whose sides are parallel to the planes
$xy$, $yz$ and $xz$ respectively
(Fig.~\ref{figure:Orthoglide_Workspace_Isotropic_Configuration}).

We will now describe a method for determining  a cube that is
enclosed in this workspace whose edge length is $2R$ such that
there is no other cube enclosed in the workspace with an edge
length of $2(R+\alpha)$, where $\alpha$ is an accuracy threshold
fixed in advance.

The first step is to determine the largest cube enclosed in the
workspace with a center located at (0,0,0). This is done by using
the ${\cal M}$ module on the box $[-A-k\alpha,A+k\alpha],
[-A-k\alpha,A+k\alpha], [-A-k\alpha,A+k\alpha]$ where $k$ is an
integer initialized to 1 and $A$ a number initialized to 0. Each
time the ${\cal M}$ module returns 1 (which means that the cube with
edge length $2(A+k\alpha)$ is enclosed in the dextrous workspace)
 we set $A$ to $A+k\alpha$ and
we double the value of $k$. When this module returns -1 if $k$ is
greater than 1 we reset $k$ to 1 and restart the process, otherwise
the process stops and we have determined that the cube with edge
length $2A$ is enclosed in the workspace while the cube with edge
length $2(A+\alpha)$ is not: thus $A$ is an initial value for $R$.

We use then the same algorithm than for the determination of the
workspace with the following modifications:
\begin{itemize}
\item  for each box $B_i$ in ${\cal L}$
we test if at the 8 corners of the cube centered at the center of
the box and with edge length $2(R+\alpha)$ the eigenvalues
satisfies the constraints:
       \begin{itemize}
       \item if this not the case let $u$ be the maximal half-width of the ranges of $B_i$
and assume that $u_1=R-u >0$. If for at least for one corner of
the cube ${\cal C}$ with center of $B_i$ and edge length $u_1$ the
eigenvalues do not satisfy the constraint, then the center of the
largest cube cannot be located in $B_i$: indeed any cube with a
center in $B_i$ and edge length $2R$ will include all the corners
of ${\cal C}$ and thus contain at least a point for which the
eigenvalues do not satisfy the constraints.
        \item if at the corners the eigenvalues are all valid then we
search for the largest cube centered at the center of the box
using the same method than for the determination of the largest
cube centered at (0,0,0). If we find a cube with a larger edge
length than $2R$, then the value of $R$ is updated.
        \end{itemize}
\item the boxes are bisected only if the largest width of their ranges
is greater than $2\alpha$: indeed even if the largest cube has a
center located in such box, then the maximal edge length will be
at most $2(R+\alpha)$.
\end{itemize}
Using this algorithm with a value of 0.001 for $\alpha$ we found
out that the largest cube has its center located at $(0.085938,
0.085938, 0.085938)$ and that its edge length was $L_{Workspace}=
0.643950$ while we may guarantee that there is no cube with edge
length larger than $0.643952$ enclosed in the workspace
\section{Conclusions}
The dextrous Cartesian workspace and the largest cube enclosed in
this workspace is computed using interval analysis based method
for the Orthoglide. Unlike most existing PKMs, the dextrous
workspace is fairly regular and the performances are homogeneous
in it. Thus, the entire workspace is really available for tool
paths. The bounds of velocity and force transmission factors used
in this paper are given as an example because it depends on the
performance requirements for machining applications. A 1:3-scale
prototype of this mechanism is under construction in our
laboratory with these bound contraints.

\begin{chapthebibliography}{1}
\bibitem[Treib and Zirn, 1998]{Treib} 
Treib, T.\ and Zirn, O.\, (1998),
\newblock ``Similarity laws of serial and parallel
manipulators for machine tools'',
\newblock Proc. Int. Seminar on Improving Machine Tool Performance, pp.~125--131, Vol.~1.
\bibitem[Wenger et al., 1999]{Wenger_a} 
Wenger, P.\, Gosselin, C.\ and Maille. B.\, (1999),
\newblock ``A Comparative Study
of Serial and Parallel Mechanism Topologies for Machine Tools'',
\newblock Proc. PKM'99, Milano, pp.~23--32.
\bibitem[Kim et al., 1997]{Kim} 
Kim J. , Park C., Kim J. and Park F.C., (1997), ``Performance
Analysis of Parallel Manipulator Architectures for CNC Machining
Applications'', Proc. IMECE Symp. On Machine Tools, Dallas.
\bibitem[Wenger et al., 2001]{Wenger_b} 
Wenger, P.\, Gosselin, C.\ and Chablat, D.\, (2001),
\newblock ``A Comparative Study
of Parallel Kinematic Architectures for Machining Applications'',
\newblock Proc. Workshop on Computational Kinematics', Seoul, Korea, pp.~249--258.
\bibitem[Rehsteiner et al., 1999]{Rehsteiner_a} 
Rehsteiner, F., Neugebauer, R. Spiewak, S. and Wieland, F.,
(1999),
\newblock ``Putting Parallel Kinematics Machines (PKM) to Productive Work'',
\newblock Annals of the CIRP, Vol.~48:1, pp.~345--350.
\bibitem[Rehsteiner  et al., 1999]{Rehsteiner_b} 
Tlusty, J., Ziegert, J, and Ridgeway, S., (1999),
\newblock ``Fundamental Comparison of the Use of Serial and Parallel Kinematics for Machine Tools'',
\newblock Annals of the CIRP, Vol.~48:1, pp.~351--356.
\bibitem[Luh  et al., 1996]{Luh} 
Luh  C-M., Adkins F. A., Haug E. J. and Qui C. C., (1996),
``Working Capability Analysis of Stewart platforms'', Transactions
of ASME, pp.~220--227.
\bibitem[Merlet, 1999]{Merlet} 
Merlet J-P., (1999), ``Determination of 6D Workspace of Gough-Type
Parallel Manipulator and Comparison between Different
Geometries'', The Int. Journal of Robotic Research, Vol.~19,
No.~9, pp.~902--916.
\bibitem[Golub and Van Loan, 1989]{Golub}
Golub, G. H. and Van Loan, C. F., (1989), Matrix Computations, The
John Hopkins University Press, Baltimore.
\bibitem[Salisbury and Craig, 1982]{Salisbury}
Salisbury J-K. and Craig J-J., (1982), ``Articulated Hands: Force
Control and Kinematic Issues'', The Int. J. Robotics Res., Vol.~1,
No.~1, pp.~4--17.
\bibitem[Angeles, 1997]{Angeles}
Angeles J., (1997), Fundamentals of Robotic Mechanical Systems,
Springer-Verlag.
\bibitem[Wenger and Chablat, 2001]{Wenger} 
Wenger, P., and Chablat, D., (2000), ``Kinematic Analysis of a new
Parallel Machine Tool: the Orthoglide'', in Lenar\v{c}i\v{c}, J.
and Stani\v{s}i\'c, M.M. (editors), {\em Advances in Robot
Kinematic}, Kluwer Academic Publishers, June, pp.~305--314.
\bibitem[Chablat and Wenger, 1998]{Chablat} 
Chablat D. and Wenger P., (1998), ``Working Modes and Aspects in
Fully-Parallel Manipulator'', IEEE Int. Conf. On Robotics and
Automation, pp.~1964--1969.
\bibitem[Merlet, 2000]{Merlett} 
Merlet J-P., (2000), ``{ALIAS}: an interval analysis based library
for solving and analyzing system of equations'', SEA, June.
Automation, pp.~1964--1969.
\bibitem[Clavel, (1990)]{Clavel:1990}
Clavel, R., (1990), ``Device for the Movement and Positioning of
an Element in Space", US Patent No. 4,976,582, December 11.
\bibitem[Ottaviano, (1991)]{Ottaviano}
Ottaviano E., Ceccarelli M., (2001), ``Optimal Design of CAPAMAN
(Cassino Parallel Manipulator) with Prescribed Workspace", 2nd
Workshop on Computational Kinematics CK 2001, Seoul.
pp.35-44.\end{chapthebibliography}
\end{document}